\documentclass[10pt,twocolumn,letterpaper]{article}

\usepackage{iccv}
\usepackage{times}
\usepackage{epsfig}
\usepackage{graphicx}
\usepackage{amsmath}
\usepackage{amssymb}
\usepackage{amsmath}
\usepackage{afterpage} 
\usepackage[font=small]{caption}
\usepackage[flushleft]{threeparttable}
\usepackage{array,booktabs,makecell}
\newcolumntype{L}{>{\centering\arraybackslash}m{3cm}}
\usepackage{cite}

\usepackage[breaklinks=true,bookmarks=false]{hyperref}

\iccvfinalcopy 

\begin{document}

\title{Skeleton based Zero Shot Action Recognition in Joint Pose-Language Semantic Space}

\author{Bhavan Jasani\\
Robotics Institute\\
Carnegie Mellon University\\
{\tt\small bjasani@alumni.andrew.cmu.edu}
\and
Afshaan Mazagonwalla\\ 
Electrical and Computer Engineering\\
Carnegie Mellon University\\
{\tt\small amazagon@alumni.andrew.cmu.edu}
}

\maketitle

\begin{abstract}

How does one represent an action ? How does one describe an action that we have never seen before ? Such questions are addressed by the Zero Shot Learning paradigm, where a model is trained on only a subset of classes and is evaluated on its ability to correctly classify an example from a class it has never seen before. In this work, we present a body pose based zero shot action recognition network and demonstrate its performance on the NTU RGB-D dataset. Our model learns to jointly encapsulate visual similarities based on pose features of the action performer as well as similarities in the natural language descriptions of the unseen action class names.  We demonstrate how this pose-language semantic space encodes knowledge which allows our model to correctly predict actions not seen during training.

\end{abstract}

\section{Introduction}

Most of the current approaches for action recognition require large well labelled datasets and work only on the action classes the model is trained on. New datasets for action recognition are frequently being released, and the number of action categories keeps on increasing. The action categories may thus be extremely fine grained and the models work well on the predefined action categories but fail to generalize when given an example of a category outside of the training set. At the same time it is difficult to get sufficient training data for new actions. Lack of training examples of certain categories is a common problem in computer vision and to tackle this a lot of work has been done in the area of Zero Shot Learning.

In Zero-Shot Learning (ZSL) the visual model is trained with visual data from a subset of the available classes called the seen classes and it subsequently learns to generalize to previously unseen classes with the help of some external information contained in some other modality or sources of data that are easily available. A majority of previous work in zero-shot learning use text in the form of word embeddings as the external knowledge base, as unannotated text data is easily available. This is similar to the way humans can indirectly learn about novel things just by reading the description of an image or video without visually looking at them along with the knowledge we already have about similar things we have seen. Significant amount of prior work \cite{DeViSE,LTC,LTC_SSE} in zero shot learning have focused on image classification by learning joint space between the visual features from RGB images and the text embeddings of the class labels. Recently these ideas have been started to extend to action recognition.

Earlier work such as \cite{action_PLD} have shown that actions can be recognized from point light displays of body pose, opening an exciting area of body pose based action recognition \cite{stgcn2018aaai}. For these models the actions are recognized from spatio-temporal dynamics of human body joints of the action performer rather than using the full RGB images/videos. Further, recent deep learning based human body pose detectors like OpenPose \cite{openpose} work really well in diverse conditions to detect body pose. \cite{he2016human} has shown that at times the visual background can be strong enough to correctly predict the action class even when the human is not present in the video, which ideally should not happen. This motivates us to use visual features derived from the skeletal pose of a human which captures the most salient motion associated with an action without being biased by the visual background or context, thereby allowing it to generalize across variations in different environments as well as other potential visual artefacts like lighting, contrast, or in an extreme case, adversarial perturbations.

We are interested in combining these two concepts - zero shot learning from text embeddings and pose based action recognition to learn a joint semantic space of pose and language, and explore if this can allow models to correctly predict novel unseen actions. For example, if the model has been trained on examples of a person \textit{"wearing a jacket"} (based on body pose only). Then is it possible to correctly infer a similar looking action which is not present in the training set, like a person \textit{"removing the jacket"} based on the body pose? This is a harder problem than it would seem, since for extremely related classes, a naive model trained would always predict the class available during training, whereas, a good model should be able to label \textit{"removing the jacket"} as its own class in the transductive or Generalized Zero Shot Setting.

In this paper, we demonstrate zero shot learning using pose-language semantic space on NTU RGB-D dataset and hope this opens up avenues for further research in the field. This paper is divided into different sections. In section 2 describe related work, in section 3 we describe our model, in section 4 we describe the experimental details of the dataset and model and show the results in section 5.

 \begin{figure*}[thpb]
      \centering
      \includegraphics[scale=0.7]{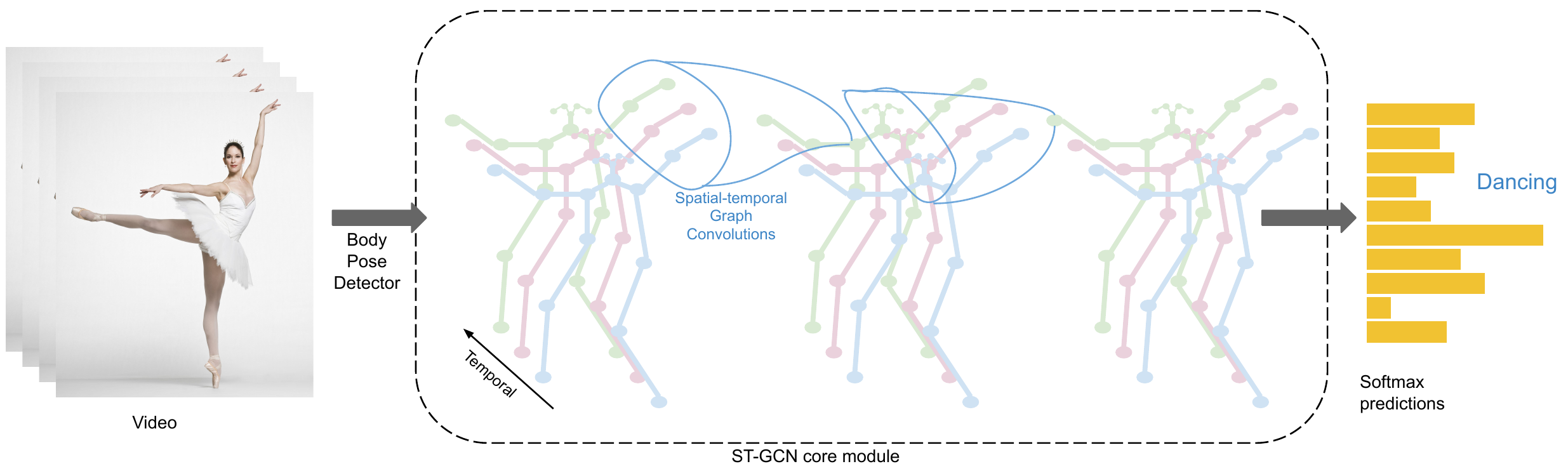}
      \caption{Spatial Temporal Graph Convolutional Network (ST-GCN) model}
      \label{figurelabel}
  \end{figure*}

\section{Related works}

\textbf{Action Recognition:} Within the action recognition community, there has been extensive research towards finding an ideal video representation that captures the salient features of a video sample in order to classify it as one of the available categories. A variety of methods have been applied to this effect including Two-stream Networks \cite{simonyan2014two} that jointly learn spatial and temporal features of videos, 3D Convolutions \cite{ji20133d} that apply a learnable convolutional kernel directly over the RGB video frames across time, optical flow based methods \cite{wang2011action} that capture frame level motion dynamics and Pose Based Methods \cite{cheron2015p} that directly model the dynamics of body keypoints.

\textbf{Pose Based Action Recognition:} There have been a studies in Human Vision \cite{bulthoff1998top,action_PLD} that also suggest that human pose can provide sufficient information for determining human actions. Also, He \etal \cite{he2016human} found that visual context can inadvertently bias the model to  predict the correct action class for Human Action Recognition based on the surrounding pixels even when the human is not present in the video.

A pose is defined a set of key points per frame denoting a fixed number of body joint locations. In Pose based action recognition, the video is represented as a sequence of poses, i.e a sequence of a set of 2D or 3D coordinates from which actions are classified. Previous methods in pose based action recognition like \cite{wang2013approach} use rule based parsing techniques to manually group body key points based on rules of human anatomy. Recent architectures such and ST-GCN \cite{stgcn2018aaai} and SR-TSL \cite{si2018skeleton} use graph convolutional networks and other end to end trainable architectures that allow for these representations to be learnt purely from the available data by jointly optimizing parameters of the representation network and the classification network.

\textbf{Zero Shot Action Recognition:} The focus of research of the zero shot recognition community is in finding the right semantic space in which to project visual features so that similar classes appear close together and dissimilar classes appear further apart. A bulk of this work is of the form where visual features are projected onto the language space. One of the early works to demonstrate this idea was DeViSE \cite{DeViSE} which uses a simple learned linear projection between the visual feature space and the class name embeddings. The model can then assign a class label to the unseen example using a fixed nearest neighbor or linear classifier.

In recent times, there have been a number of interesting approaches to solving the zero shot learning problem using Error Correcting Codes \cite{qin2017zero}, Generative Models \cite{Mishra2018AGA} and using a Learned Distance Metric \cite{LTC}. While most other works use a fixed distance metric with which to assign a nearest neighbor class to an unseen example, in Learning to Compare (LTC) \cite{LTC} the authors tackle a zero shot problem by learning the metric as an optimization problem based on the training data.
In \cite{ZSLgcn} the authors build knowledge graphs to learn relationships between action class names as well as Two Stream  Convolutional networks for learning visual features from the RGB image frame. In contrast, our paper uses skeleton based graph convolutions to encode visual features within the action video. 


\section{Approach}

Our zero shot action recognition model is modular, it consists of three components: 1) Skeleton based action recognition network 2) External language knowledge base 3) Zero shot learning network. 

We use Spatial Temporal Graph Convolution Network for skeleton based action recognition (ST-GCN) \cite{stgcn2018aaai} which acts as the visual feature extractor. We use Sent2Vec \cite{sent2vec} trained on a huge English Wikipedia text corpus to generate class label embeddings, which provides our model with external knowledge. And Learning to Compare - Relation Network \cite{LTC} is used for jointly learning the visual pose features and the external language based knowledge base. 

The total action classes are divided into seen (used for training) and unseen classes (used only for testing). The input to ST-GCN is the time series of body pose key points. We first train the ST-GCN on a subset of the available action class data (the seen classes) and then use the trained model to extract the visual features for all the seen classes training data. In parallel, we obtain the language embeddings of all the action class labels (seen + unseen) using Sent2Vec \cite{sent2vec} pre-trained on 69 million English sentences from Wikipedia texts. This encodes the external knowledge about all the classes, seen as well as unseen. Using the visual features of training examples of seen classes from ST-GCN and class label embeddings from Sent2Vec we then train a zero shot model which learns to match the visual features with the class label embeddings (that encodes external knowledge). We use Relation network \cite{LTC} for zero-shot learning and as a baseline also experiment with DeViSE \cite{DeViSE}. 

During test time, we first pass the videos of unseen action class through ST-GCN to compute it's visual features, which are then passed as inputs to Relation network or DeViSE which finds the nearest relevant class label embedding corresponding to the visual features and hence provides the output class label for the unseen action. Each part of our model are explained in greater detail in the following sections:

\begin{figure*}[t]
      \centering

    \vspace*{-0.4cm}
      \includegraphics[scale=0.9]{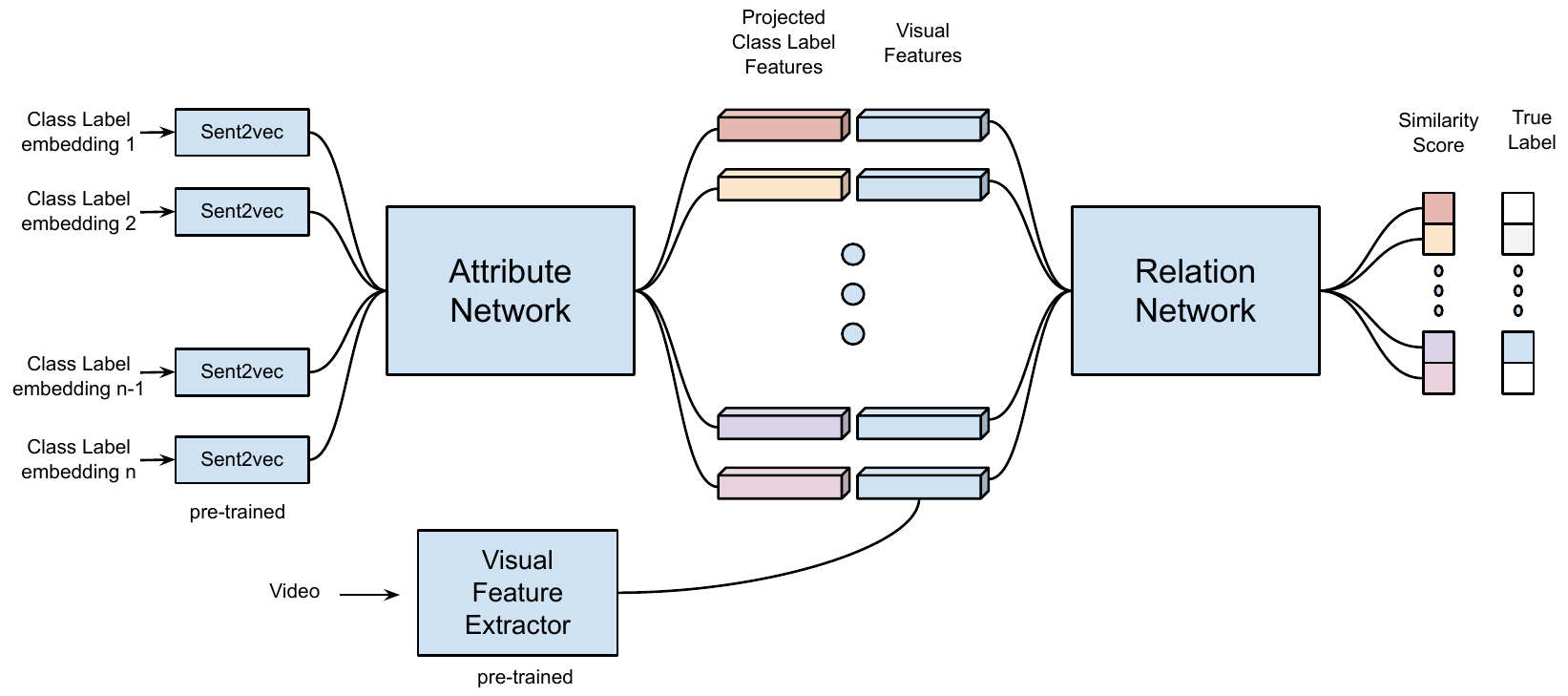}
      \caption{Relation Network Architecture for Zero-Shot Learning}
      \label{figurelabel}
   \end{figure*}


\subsection{Spatio Temporal Graph Convolutional Network}

Human actions can be recognized by the dynamics of skeleton. Since skeletons are in the form of a graph ST-GCN uses Graph Convolutions Network which is a generalization of Convolutional Neural Networks to work on graphs of arbitrary structures. Further since the input is time series of skeleton, it uses spatial-temporal graph convolutions. 

ST-GCN takes as input the time series of body pose of the action performer in the given video, which can be generated using a human body pose detector. It works with both 2D and 3D body key points. The input to the ST-GCN is a graph where each node represents a body joint, and there are two type of edges - the spatial edges resembles the natural connections in human bodies and the temporal edges connect every body joint to itself across the temporal sequence. ST-GCN then applies multiple layers of spatial-temporal convolutions on the neighbouring spatial and temporal nodes in the input graph just like CNN's. This results into hierarchical higher-level feature representations similar to CNN's. This way ST-GCN learns to hierarchically capture the spatial and temporal dynamics of human body movements. Eventually after multiple layers of graph convolutions and pooling, a soft-max layer is applied which gives probability distribution for the corresponding action categories.

\subsection{External language based knowledge base}
For most of the action recognition datasets the class labels are in the form of phrases (\eg \textit{“wearing the jacket, taking-off the jacket”}) instead of single words and so we use Sent2Vec \cite{sent2vec} for generating class label embeddings. The Sent2Vec model is an extension of the Continuous Bag of Words (CBOW) model for word contexts to a larger sentence context that uses an unsupervised objective to train distributed representations of sentences from a large corpus of text data. For our task, we generate bigram embeddings trained on a 16GB corpus of English Wikipedia texts, which contains about 69 million sentences and about 1.7 billion words \cite{sent2vec}. The resulting class label embeddings are a vector of size 700 dimension. This allows our zero-shot model to learn the meaning and language semantics of various action class labels (of both seen and unseen classes) from naturally occurring text data and therefore serves as the external knowledge base. In this embedding space, similar actions will be closer. For example \textit{walking} and \textit{running} will have higher similarity of it's embedding in comparison with that of \textit{wearing a jacket}. So even if the action recognition model has only seen visual examples of \textit{running}, based on these language embeddings it will know that \textit{walking} is similar to \textit{running}. This eventually would help in zero shot learning.

\subsection{Zero shot learning - DeViSE}

Deep Visual-Semantic Embedding Model (DeViSE) \cite{DeViSE} is one of the first deep learning based work on learning a visual-semantic model using unannotated texts for zero-shot image classification. In our work we use this for zero-shot action recognition instead of image classification, and use it as a baseline. DeViSE uses a projection matrix (a linear transformation) between the visual features and the class label embedding, so as to project the visual features in the space of class label embeddings.

For learning the projection matrix the DeViSE model uses a loss function which is combination of hinge rank loss and dot-product similarity. The objective here is to produce higher dot-product similarity between the output of visual model and the class embedding of the correct class and lower dot-product similarity for the class label embeddings of all other classes.

\begin{equation}
\begin{split}
 & loss(image,label) = \\ &\sum_{j \ne label} max[0,margin - \vec{t}_{label}M\vec{v}(image) +\vec{t}_{j}M\vec{v}(image)]
\end{split}
\end{equation}

In the loss equation above $\vec{v}(image)$ is a column vector denoting the visual feature of the given video during training, $M$ is the linear projection matrix with trainable parameters and  $\vec{t}_{label}$ is a row vector of class label embedding of the true class, and $\vec{t}_{j}$ for all other classes (seen and unseen). During test time, when a video of an unseen class is given, first we find its visual representation and then project it into the space of class label embeddings using the DeViSE projection matrix and then find the class label embedding with largest dot product similarity and use it's label as the prediction of the model.

\subsection{Zero shot learning - Relation Networks}

Learning to Compare: Relation Network for Few-Shot Learning \cite{LTC} overcomes some of the limitations of DeViSE\cite{DeViSE}. The model is originally formulated for few shot learning but it generalizes to zero short learning.

It consists of 2 networks - the attribute network and the relation network. The attribute network takes as input class label embeddings and projects them to the space of visual features, it is trained to learn a good projection space. The projected class label embedding and the visual feature are concatenated depth wise and passed to the relation network. The relation network outputs similarity score for the projected class label embeddings and the visual features. Mean squared error is used as the loss, the true label being the similarity score which is 1 for the correct pair of visual features and class label embeddings and 0 otherwise. It is trained to learn a good non-linear distance metric. An episode based training strategy is used, in every episode some visual features are sampled from the training set (seen classes) and are compared with all the class label embeddings (seen + unseen classes).

In contrast to DeViSE, the Relation Network can be seen as both learning a good projection and learning a deep non linear metric (similarity function). The advantage of using this approach is that fixed metrics like in DeViSE are critically dependent on the quality of learned embedding, and they are limited by the extent to which the semantic space can generate adequately discriminative representations. In contrast, by deep learning a nonlinear similarity metric jointly with the projection function, Relation Network can better identify matching and mismatching pairs. Thereby  better co-relate the visual features with the external knowledge from class label embeddings.

\section{Experiments}

\subsection{Dataset}

NTU RGB+D\cite{NTU} is a large scale dataset for 3-D human activity analysis in an indoor environment. It consists of RGB videos, depth maps, skeleton sequences and infrared frames collected with Microsoft Kinect 2. In total, it provides 56,000 videos for 60 different action classes of daily, health-related and mutual (involving 2 people interacting) actions with 40 distinct subjects recorded from three different camera viewpoints. Each video is annotated with 3D joint locations (X, Y, Z) of 25 body key-points per subject for at most 2 subjects.


\begin{table*}[!t]
  \captionsetup{justification=centering}
  \caption{Results for Zero Shot Setting\\ (Predict on unseen classes only - Top 1 Performance)} 
  \centering 
  \begin{threeparttable}
    \begin{tabular}{c|L|ccc}
      & & & Top 1 Accuracy (\%) &  \\
    Model & Language Embedding Type &  Nearest Split  & Random Split & Furthest Split\\
     \midrule\midrule
    DeViSE & Learned  &   \textbf{75.16} & \textbf{68.47} &  42.06  \\
    \hspace{1cm} & Random    &   15.08 & 27.95 &   17.27       
    \\
    \cmidrule(l  r ){1-5}
     Relation Net & Learned  & 74.5 & 65.53 & \textbf{50.06} \\ 
     \hspace{2 cm}& Random & 11.65 & 44.88 & 3.79\\
    \midrule\midrule
    \end{tabular}
    \begin{tablenotes}
\item{DeViSE performs marginally better than Relation Net for Nearest and Furthest Split, but we see a significant difference in accuracy for the furthest split where Relation Net outperforms DeViSE indicating better separability between unrelated classes }
\end{tablenotes}

\end{threeparttable}
  \end{table*}
  


\subsection{Selecting Appropriate Class Splits}
 
In order to evaluate our model, we split the 60 available action classes in NTU-RGB+D into 55 seen training classes and 5 unseen test classes. During the training phase the model learns only from the 55 seen classes. Since there are multiple possible combinations $(^{60}C_{5})$ to pick these 5 unseen classes and there is no fixed criteria in particular for this dataset for zero shot learning, we came up with a way which divides the splits based on difficulty levels - from the easiest to the most difficult.

\textbf{Nearest Split:} Heuristically speaking this should be the easiest split. In this setting, we select our unseen classes based on the availability of a very similar class in the training set. We find the nearest neighbours of of all 60 classes based on their language embeddings. And then we pick the top few classes with least distance from other classes as our unseen classes, ensuring at the same time a sufficient amount of inter class variation between unseen test class names.

\textbf{Furthest Split:} Heuristically speaking this should be the toughest split. In this setting, we select our unseen classes which are the most dissimilar to the seen class in the training set. We find the furthest neighbours of of all 60 classes based on their language embeddings. And then we pick the top few classes with highest distance from all other classes as our unseen classes. This split is a true test of the generalization capabilities of our semantic space, as the model needs to learn very strong semantics in order to perform well on this split.

\textbf{Random Split:} In this setting, the unseen action classes are selected randomly from the available action classes. We expect this split to be of intermediate difficulty.


\subsection{Details Of The Model}

\textbf{ST-GCN} takes as input a 300 frame sequence of 25 3-D body joint locations (X,Y,Z) of 2 most prominent people in the video, and eventually computes the softmax classification probabilities of the 60 action classes used in NTU-RGB+D dataset. We train it with similar settings as that of the original model released by the authors on the seen classes for 80 epochs with SGD with base learning rate = 0.01, weight decay = 0.0001 and batch size = 48. We extract the 256 dimensional visual features from ST-GCN just before the final average pooling and softmax layer.

\textbf{For DeViSE}, the projection matrix is represented as a fully connected layer which takes as input the 256-D visual feature and projects it to 700-D vector, the size of our language embeddings. We use the hinge margin value = 0.1 in the loss calculations, and train it for 100 epochs with SGD with learning rate = 0.001, momentum = 0.9, and batch size = 64.

\textbf{The Relation Network} model, consists of two separate neural networks - attribute net and relation net. The attribute net consists of 2 fully connected layers (ReLU's are used as activation function) which takes as inputs the language embeddings and projects them to the dimensions of visual features. The relation net also consists of two fully connected layer, it takes as input the concatenation of the visual features and the output of attribute net (projected language embeddings) and outputs a single number (Sigmoid is applied at the output) which is the relation score between the visual and the language embedding. We train both the networks for about 400000 episodes sampling from batch of 32 and use ADAM as the optimizer starting with a learning rate = 1e-5 and decay it with a step size = 200000 and gamma = 0.5. For all our experiments we normalize the visual and language features to be of unit norm.


\subsection{Zero Shot Testing Paradigm}

We test our implementations based on the two zero shot learning test paradigms. In the ZSL paradigm, the set of classes available during training and the ones used for testing are disjoint. i.e, none of the classes seen during training are used for evaluating the performance during test time. While in GZSL (Generalised Zero-Shot Learning) paradigm, the model is evaluated on all the classes which is a more challenging setting. We compute “flat” hit@k metrics $–$ the percentage of test samples for which the model returns the one true label in its top k predictions, for k=1 and k=5.

\begin{table*}[t]
  \captionsetup{justification=centering}
  \caption{Results for Generalized Zero Shot Setting\\ (Predict on both seen and unseen classes - Top 1 \& Top 5 Performance)} 
  \centering 
  \begin{threeparttable}
    \begin{tabular}{c|L|ccc}
      & & & Top 1 / Top 5 Accuracy (\%) &  \\
    Model & Language Embedding Type &  Nearest Split  & Random Split & Furthest Split\\
     \midrule\midrule
    DeViSE & Learned  &  0.00 / 12.73 & 0.02 / 9.84 & 0.00 / 0.00  \\
    \hspace{1cm} & Random    &  0.00 / 0.00 & 0.00 / 0.00 &   0.00 / 0.00      
    \\
    \cmidrule(l  r ){1-5}
     Relation Net & Learned  & \textbf{19.32 / 43.19} & \textbf{20.16 / 47.14} & \textbf{14.45 / 45.49} \\ 
     \hspace{2 cm}& Random & 0.00 / 0.00 & 0.00 / 1.83 & 0.05 / 1.33\\
    \midrule\midrule
    \end{tabular}
    \begin{tablenotes}
\item{The performance of our models for Top 1 \& 5 accuracy in the Generalized Zero shot setting. Relation Network significantly outperforms DeViSE in this harder setting.}

\end{tablenotes}

\end{threeparttable}
  \end{table*}

\section{Results}

We report the performance of our baseline zero-shot learning model which uses DeViSE and of our proposed model based on Relation Network for the three different splits we described. For each of these cases, we report top1 accuracy for unseen examples evaluated over just the 5 unseen classes (ZSL performance) and unseen examples evaluated over all the 60 classes (GZSL performance)

\subsection{Performance of Vanilla ST-GCN}

As expected, vanilla ST-GCN without the zero shot learning module trained only the seen classes, but tested on the unseen classes gives 0 accuracy. When one adds the zero shot learning modules (DeViSE and Relation Network), then model is able to predict correctly the unseen classes on which it's never trained as can been seen the two tables. This thus indicates the pose-language space provides an exciting possibility for zero shot learning.

\subsection{Comparison between DeViSE and Relation Network }
From the tables it is evident that the DeViSE baseline model does a little bit better than Relation Network based model in the ZSL setting when the predictions are to be made only from the unseen classes. For GZSL setting, our proposed Relation Network based model outperforms DeViSE based baseline by a far huge margin for both top1 and top 5 accuracy, this indicates that Relation Network is able to learn the semantic relationship between the visual features and semantic features and provides better separability between unrelated classes. DeViSE on the other hand can just distinguish amongst the unseen classes but not when all possible classes are present. This emphasizes importance of having a learn-able comparison metric as implemented in Relation Network model.   

\subsection{Importance of External Knowledge}
To find the importance of external knowledge, we use randomly generated embeddings, i.e. use same models and same training settings but just use some randomly generated embeddings of the class (instead of pre-trained from text corpus). This demonstrates the zero shot performance increase due to the language semantics when the actual embeddings are used. The two tables show how using external knowledge helps our model in zero-shot learning. Without using pre-trained language embedding model, the performance of our action-recognition is close to random chance or even worse when given unseen videos. And hence shows how using external knowledge from text corpus helps our models to correctly predict actions of for the classes on which it wasn't trained.

\subsection{Comparison between different splits }
As expected in general both the models show best performance in the Nearest neighbour split and the worst performance in the Furthest neighbour split, and intermediate performance in the randomly selected split. This shows the best and worst limits of our zero shot models.

\section{Conclusion}

We demonstrate a pose based zero shot action recognition framework which uses spatial-temporal graph convolutions to generate visual features and along with class label embeddings generated from pre-trained Wikipedia text corpus can predict novel actions not seen during training time. And we focus on using pose only information of the action performer without looking at context for zero shot learning. Based on our results the use of learnable similarity metric for learning the similarity between visual features and class label embedding contributes significantly to the classification accuracy of the model for examples from unseen classes.

{\small
\bibliographystyle{ieee}
\bibliography{egbib}
}

\end{document}